\documentclass[10pt,journal,compsoc]{IEEEtran}

\ifCLASSOPTIONcompsoc
  \usepackage[nocompress]{cite}
\else
  \usepackage{cite}
\fi

\ifCLASSINFOpdf

\else

\fi

\usepackage{graphicx,comment}
\usepackage{amsmath}
\usepackage{siunitx}
\usepackage[utf8]{inputenc} 
\usepackage{amsfonts}       
\usepackage{nicefrac}       
\usepackage{microtype}      
\usepackage{amsmath, amsfonts,bm,amssymb,indentfirst,mathtools,bbm,mathtools}
\usepackage[noend]{algpseudocode}
\usepackage{algorithmicx,algorithm,enumerate,accents}
\usepackage{color}
\usepackage[hidelinks]{hyperref} 

\allowdisplaybreaks[4]

\usepackage{amsthm,amsmath,amsfonts,amssymb}
\usepackage{amsmath}
\usepackage{comment}
\usepackage{booktabs}       
\usepackage{amsfonts}       
\usepackage{nicefrac}       
\usepackage[noend]{algpseudocode}
\usepackage{amsmath}
\usepackage{lastpage}
\usepackage{enumerate}
\usepackage{fancyhdr}
\usepackage{mathrsfs}
\usepackage[table]{xcolor}
\usepackage{listings}
\usepackage{bbm}
\usepackage{subfig}
\usepackage{caption}
\usepackage{diagbox} 
\usepackage[switch]{lineno}

\usepackage{tikz-qtree,tikz-qtree-compat}
\usepackage{tikz,pgf,hyperref}
\usetikzlibrary{positioning,patterns}
\usetikzlibrary{calc,fadings,decorations.pathreplacing,arrows,
decorations.markings,
datavisualization.formats.functions,shapes.geometric}
\usetikzlibrary{hobby}
\usetikzlibrary{shadows}

\def\boxit#1{\vbox{\hrule\hbox{\vrule\kern6pt
          \vbox{\kern6pt#1\kern6pt}\kern6pt\vrule}\hrule}}

\def\cov{\hbox{cov}}

\def\bse{\begin{eqnarray*}}
\def\ese{\end{eqnarray*}}
\def\be{\begin{eqnarray}}
\def\ee{\end{eqnarray}}
\def\bq{\begin{equation}}
\def\eq{\end{equation}}
\def\bse{\begin{eqnarray*}}
\def\ese{\end{eqnarray*}}

\def\T{^{\rm T}}

\newcommand{\jrev}[1]{\textcolor{black}{#1}}
\newcommand{\wrev}[1]{\textcolor{black}{#1}}

\newcommand{\crb}[1]{\textcolor{black}{#1}}

\renewcommand{\baselinestretch}{1} 

\newcommand{\bbS}{\mathbb{S}}

\newcommand{\bL}{\mathbf{L}}

\newcommand{\bW}{\mathbf{W}}
\newcommand{\bP}{\mathbf{P}}

\newcommand{\bC}{\mathbf{C}}
\newcommand{\bA}{\mathbf{A}}

\newcommand{\bS}{\mathbf{S}}

\newcommand{\bX}{\mathbf{X}}
\newcommand{\bY}{\mathbf{Y}}

\newcommand{\bx}{\mathbf{x}}
\newcommand{\by}{\mathbf{y}}

\newcommand{\ba}{\mathbf{a}}
\newcommand{\bb}{\mathbf{b}}

\newcommand{\bc}{\mathbf{c}}

\newcommand{\bu}{\mathbf{u}}

\newcommand{\bk}{\mathbf{k}}

\newcommand{\bgamma}{\boldsymbol{\gamma}}

\newcommand{\blambda}{\boldsymbol{\lambda}}

\newcommand{\btheta}{\boldsymbol{\theta}}

\newcommand{\bvarepsilon}{\boldsymbol{\varepsilon}}

\newcommand{\bbeta}{\boldsymbol{\beta}}

\newcommand{\0}{\mathbf{0}}

\newcommand{\mcN}{{\mathcal N}}
\newcommand{\mcO}{{\mathcal O}}
\newcommand{\mcP}{{\mathcal P}}

\newcommand{\mcS}{\mathcal{S}}

\newcommand{\eset}[1]{{\mathbb E} \left[ #1 \right] }

\def\R{\Bbb{R}}

\theoremstyle{plain}

\newtheorem{theorem}{Theorem}[section]

\theoremstyle{remark}

\newtheorem*{remark}{Remark}

\tikzstyle{block} = [draw, shade, drop shadow, rounded corners=1ex,
bottom color= 
green!40!,
minimum width=2cm] 

            \tikzstyle{vecArrow} = [thick, blue,,
            decoration={markings,mark=at position 1 with
              {\arrow[semithick,blue]{open triangle 60}}}, double
            distance=1.4pt, shorten >= 5.5pt, preaction = {decorate},
            postaction = {draw,line width=1.4pt, white,shorten >=
              4.5pt}]

\begin{document}
%
\title{Multilevel Stochastic Optimization for Imputation in Massive Medical Data Records}
%
%
%
%

\author{Wenrui Li,~
        Xiaoyu Wang,
        Yuetian Sun,
        Snezana Milanovic,
        Mark Kon,~
        Julio Enrique Castrill\'on-Cand\'as
\IEEEcompsocitemizethanks{\IEEEcompsocthanksitem Snezana Milanovic is with Sunovion Pharmaceuticals, Marlborough MA, 01752. All other authors are with the Department of Mathematics \& Statistics, Boston University, Boston MA, 02215.\protect\\
E-mail: jcandas@bu.edu
}
}

\IEEEtitleabstractindextext{%
\begin{abstract}
\jrev{It has long been a recognized problem that many datasets contain
significant levels of missing numerical data. A potentially critical
predicate for application of machine learning methods to datasets
involves addressing this problem. However, this is a 
challenging task. In this paper, we apply a
recently developed multi-level stochastic optimization approach to the
problem of imputation in massive medical records. The approach is
based on computational applied mathematics techniques and is highly
accurate. In particular, for the Best Linear Unbiased Predictor (BLUP)
this multi-level formulation is \emph{exact}, and is 
significantly faster and more numerically stable. 
This permits
practical application of Kriging methods to data imputation problems
for massive datasets. 
We test this approach on data from the National
Inpatient Sample (NIS) data records, Healthcare Cost and Utilization
Project (HCUP), Agency for Healthcare Research and Quality. Numerical
results show that the multi-level method significantly outperforms current
approaches and is numerically robust. It has superior
accuracy as compared with methods recommended in the recent report
from HCUP. 
Benchmark tests show up to 75\% reductions in error. Furthermore, 
the results are also superior to recent state of the art methods such as discriminative deep learning.} 

\end{abstract}

\begin{IEEEkeywords}
Massive Datasets, Machine Learning, Best Linear Unbiased Predictor, Computational Applied Mathematics, Numerical Stability
\end{IEEEkeywords}}

\maketitle

\IEEEdisplaynontitleabstractindextext

%
\IEEEpeerreviewmaketitle

\ifCLASSOPTIONcompsoc
\IEEEraisesectionheading{\section{Introduction}\label{sec:introduction}}
\else
\section{Introduction}
\label{sec:introduction}
\fi

\IEEEPARstart{W}{ith}
the growing emphasis on massive datasets in many modern
applications, the need for sophisticated and precise approaches to high
dimensional and heterogeneous data analysis is increasing.  As an
example, in healthcare research and personalized medicine, many
Electronic Medical Records (EMR) include data on millions of patients,
harboring large numbers of variables (e.g., demographics,
diagnostic/procedure codes, lab/imaging results).  These massive
biomedical datasets, among others, provide opportunities to advance
clinical and biomedical research, including clinical phenotyping
(i.e., learning clinical trait-related features), but such analyses
require shifting from human-guided solutions toward machine-learning
(ML)-driven approaches. ML can increase clinical prediction accuracy
and contribute to clinical phenotyping.  However, many of these
datasets are incomplete and include significant components of missing
data.

ML algorithms cannot function without complete data matrices. Removing
or imputing missing data can reduce sample sizes or bias outcomes. A
critical foundational element for studying large datasets includes
properly addressing the problem of missing 
and incorrect data (\cite{Roderick2014}).

\wrev{There is extensive work on statistical and ML methods for data imputation. These
are seen in two categories. The first category involves constructing
statistical and/or deterministic  models for specific types of datasets, including single-cell RNA-sequencing data \cite{xu2020scigans,li2018accurate,Arisdakessian2019}, image data \cite{lee2019collagan}, time series data \cite{luo2018multivariate,cao2018brits}, 
and traffic data \cite{duan2016efficient,chen2019bayesian}.
The second category includes general methods that apply to a large
class of datasets. Popular methods include k-nearest neighbors \cite{batista2003analysis}, 
discriminative deep learning methods\cite{biessmann2018deep}, and generative deep learning methods\cite{shang2017vigan,yoon2018gain,limisgan,nazabal2020handling,qiu2020genomic}.
See, for example, \cite{jager2021benchmark,emmanuel2021survey,thomas2021systematic,
    little2019statistical} for surveys of this area. 
}

Many ad hoc techniques have been developed to deal with this problem,
including sample deletion, mean value or nearest neighbor
imputation, etc.; in general these suffer from information loss that
leads to inaccurate predictions.

Missing data form an important problem in medical record datasets. 
In particular, the HCUP Report \#2015-01  by \cite{Houchens2015}, 
\emph{stresses} the need 
to address missing data in the National Inpatient Sample (NIS) and State
Inpatient Databases (SID).  As an example, suppose that there are missing 
data for discharges (total charge) in rural hospitals. Such missing data can lead to
erroneous estimates of total charges, potentially biasing or otherwise misdirecting  
state/federal funding policies. It is important to obtain accurate 
and unbiased estimates of missing data.

\jrev{In the HCUP Report \#2015-01 the missing data is spread out to the whole 
dataframe. For example, about 20\% of the data is missing for the important
total charge variable for the Michigan SID dataset. In this paper we impute the missing
values from a column of the dataframe by using the rest of the information including
other columns. For example, for the total charge variable we use the data available for total charge, length of stay, number of procedures, number of diagnoses and age.}


Current imputation algorithms recommended by the
HCUP report \#2015-01
include Predicted Mean
Matching (PMM), Predicted Posterior Distribution (PPD) and linear
regression (\cite{White2011,Schafer1999}).
These algorithms often are
sub-optimal, in particular for noisy signals. Furthermore, Bayesian
methods such as Data Augmentation (DA) see \cite{Schafer1997}) and Bootstrapping
Expectation Maximization (BEM) algorithms (see \cite{Honaker2011}) suffer
from poor accuracy. \jrev{We note that a recent approach by \cite{Bertsimas2018} improves on
  the accuracy of traditional methods such as PMM, PPD, BEM, etc, by
  using a so-called optimization layer. 
  More modern state of the art methods such as Discrimitative Deep Learning (DDL) \cite{biessmann2018deep}  produces excellent results. This has been quantified in the recent benchmark paper \cite{jager2021benchmark}}. 

\jrev{In this paper we apply stochastic optimization approaches,
such as Kriging/Best Linear
Unbiased Predictor (BLUP) \cite{Nielsen2002}.  We note that we refer to Kriging as both the estimation of the coefficients of the covariance function and BLUP, although we mostly use Kriging/BLUP for clarification 
purposes.  The technique is based on a principled
optimal probabilistic representation of the data.  These methods
  can lead to optimized imputation by taking advantage of essentially 
all available data.   Kriging is a popular method
for imputation for Geostatistics \cite{Shtiliyanova2017} and has been 
extended to other applications such as traffic flow \cite{Yang2018} due
to its high accuracy it is popular in many fields.
However, Kriging methods in their general
application are often costly and unstable numerically on massive
datasets.  A common technique is to apply a nugget to the
  covariance matrix. However, this changes the covariance function model and
  does not solve the original problem. 
This has been a limiting factor for application of
Kriging to imputation for massive datasets outside of the
spatio-temporal domain.}

One of the goals of this paper is to motivate the application of
Computational Applied Mathematical (CAM) techniques to solve large
scale stochastic optimization problems.  In particular, to
  address the above challenges, we propose to apply the recently
  developed multi-level Kriging approach that is designed to tackle
  computing cost effectiveness and numerical instability
  (\cite{Castrillon2013,Castrillon2016a,Castrillon2020}). These
techniques originate from the fields of numerical analysis and
uncertainty quantification
(\cite{Castrillon2016,Castrillon2021a,Castrillon2021b,
abramowitz1964,Castrillon2020,nobile_tempone_08,babusk_nobile_temp_10})
and have been effective in solving (stochastic) partial differential
equations.  Indeed, the present Kriging optimization problem has many
connections to the solution of Partial Differential Equations (PDEs).
We introduce the above techniques in the context of statistical methods
including Kriging, and demonstrate their power to solve hard
stochastic optimization large scale problems.

By remapping an original stochastic optimization problem onto a
multi-level space, we can significantly mitigate numerical
instabilities and reduce computational burdens. In particular, the
BLUP is remapped onto an equivalent formulation with multi-level
spaces. Mathematically the multi-level prediction is \emph{exact},
i.e., it precisely solves the original BLUP problem. In practice,
  numerical efficiency augmentations involving factors of the order of tens of thousands can
  be gained for 20 dimensional problems, as compared with traditional
  Conjugate Gradient (CG) approaches for estimates with the same
  accuracy, as is shown in Section \eqref{section:ER}.

To demonstrate the accuracy of the multi-level Kriging imputation
method, we benchmark it on the U.S. National Inpatient Sample (NIS)
datasets (see \cite{NIS2012}, Healthcare Cost and Utilization Project
(HCUP), Agency for Healthcare Research and Quality). Significant
improvements over methods including PPM, PPD, DA and
EM are shown.  More importantly, it is shown that the imputed
  values accurately reflect the overall statistics of the
  population. This contrasts with other approaches, including kNN-R,
kNN, GLS, PPM, etc, which often also suffer from poor accuracy.
\jrev{Furthermore our results show improvements over the recent Discrimitative Deep Learning state of the art method.}

%
%
%
%

\section{Problem Setup}

Suppose that for a Gaussian random field $Y$ we have the model:
\begin{equation}
Y(\bx) = \bk(\bx)\T \bbeta+\varepsilon(\bx), \qquad \bx \in \R^d,
\label{Introduction:noisemodel}
\end{equation}
where $d$ is the number of spatial dimensions, $\bk:\R^d \rightarrow
\R^p$ is a functional vector of the spatial location $\bx \in \R^{d}$,
$\bbeta\in\R^p$ is an unknown vector of coefficients. The noise model
$\varepsilon$ is a stationary Gaussian random field with mean zero and
parametric covariance function $\phi(\bx,\by;\btheta) \equiv
\cov(\varepsilon(\bx),\varepsilon(\by)):\R^{d} \times \R^{d}
\rightarrow \R$, where $\btheta\in\R^d$ is an unknown vector of
parameters. We assume $\phi(\bx,\by;\btheta)$ is positive
definite.

Suppose that we collect $N \geq p$ observations of the Gaussian random
field process $Y$ at different locations in $\R^{d}$, i.e.  the vector
of observations $\bY=(Y(\bx_1),\ldots,$ $Y(\bx_N))\T$ is
obtained from locations in the set $\bbS :=\{
\bx_{1},\dots,\bx_{N}\}$, where the elements in $\bbS$ are assumed
non-collinear.  Denote $\bC(\btheta)=\cov(\bY,\bY\T)\in \R^{N \times N}$
bas the covariance matrix of $\bY$. Furthermore,  assume  that 
$\bC(\btheta)$ is positive definite
for all $\btheta\in\R^w$.  Let $\bX=\big( \bk(\bx_1) \ldots
\bk(\bx_N)\big)\T\in \R^{N\times p}$ and assume that it is full column
rank. Since the model \eqref{Introduction:noisemodel} is a Gaussian
random field, the samples in $\bbS$ can be written in vector form
as

\begin{equation} \bY = \bX \bbeta + \bvarepsilon,
\label{Introduction:vectormodel} \end{equation}
where $\bvarepsilon$ is a Normal random vector, more precisely
$\varepsilon \sim N(0,\bC(\btheta))$. The aim is to {\it
estimate} the unknown vectors $\bbeta$ and $\btheta$ and {\it
predict} $Y(\bx_0)$ for a new spatial location $\bx_0$ with a
stochastic optimization method.  The unknown vectors $\bbeta$ and
$\btheta$ are estimated from the data using a log-likelihood
function (see \cite{Castrillon2016a}) 
\begin{align}\label{Introduction:loglikelihood}   
\begin{split}
 \ell(\bbeta,\btheta)= &-\frac{n}{2}\log(2\pi)-\frac{1}{2}\log
\det\{\bC(\btheta)\}\\
&-\frac{1}{2}(\bY-\bX\bbeta)\T\bC(\btheta)^{-1}
(\bY-\bX\bbeta),
\end{split}
\end{align}
which can be profiled by Generalized Least Squares (GLS) with
\begin{equation}
  \hat \bbeta(\btheta)=(\bX\T \bC(\btheta)^{-1} \bX)^{-1}\bX\T
  \bC(\btheta)^{-1}\bY.
  \label{GLSbeta}
\end{equation}

For the prediction problem, consider the Best Linear Unbiased
Predictor (BLUP) $\hat Y(\bx_0)=\blambda\T\bY$, where
$\blambda=(\lambda_1,\ldots,\lambda_N)\T$.  The BLUP
is formulated as the minimization of $\eset{(Y(\bx_0)-\blambda\T\bY)^2}$
under  the unbiased
constraint $\bX\T\blambda=\bk(\bx_0)$. The solution to this
problem (\cite{Nielsen2002}) is given as

\begin{equation}
\hat Y(\bx_0)=\bk(\bx_0)\T\hat \bbeta+\bc(\btheta)\T
\bC(\btheta)^{-1}(\bY-\bX\hat \bbeta), 
\label{KrigBLUP}
\end{equation}
where $\bc(\btheta)=\cov\{\bY,Y(\bx_0)\}\in \R^{N}$ and 
$\hat \bbeta$ is defined in equation \eqref{GLSbeta}.

Solving the Kriging estimation and prediction problem involves
inverting the covariance matrix $\bC(\btheta)$.  Two main approaches
exist. Direct methods, such as Gaussian elimination and Cholesky
factorizations are popular for small datasets, but as the number of
observations $N$ increases the memory constraints grow as $\mcO(N^2)$,
which makes it infeasible for large datasets. Iterative methods, such
as the Conjugate Gradient method (CG), avoid computing the covariance
matrix, thus making them a good choice for large datasets.

The key problem with direct and iterative methods is that they are
sensitive to the condition number of the covariance matrix. Large
condition numbers leads to numerical instability with the consequence
of inaccurate solutions. It can be shown that the accuracy for
\emph{any} inversion numerical algorithm is $\approx$
$\kappa(\bC(\btheta))\epsilon_{M}$ (see \cite{Jabbari}), where
$\kappa(\bC(\btheta))$ is the condition number of $\bC(\btheta)$ and
$\epsilon_M$ is the relative machine precision. For most computers,
double precision $\epsilon_{M} \approx 10^{-16}$.  Furthermore,
iterative methods such as CG are slow for matrices with large
condition numbers (see \cite{Golub1996}).  For many practical covariance
functions, the condition numbers of $\bC(\btheta)$ are large.

\textbf{Challenge:} \emph{The number of observations is large and
    the covariance matrix $\bC(\btheta)$ can be highly ill-conditioned.
    This can lead to inaccurate estimates of the
    Kriging / BLUP predictor.} In the paper written by
  \cite{Castrillon2016a}, the authors propose a new transformation of
  the data vector $\bY$, leading to a decoupled multi-level
  description of the Kriging model without any loss of structure for
  $\R^2$ and $\R^3$ problems.  As discussed in the introduction,
  missing data in the NIS and SID datasets form an important issue
  that is underscored in the HCUP report \#2015-01
  (\cite{Houchens2015}). The report shows as an example, the problem
  of missing data for discharges (total charge) in rural hospitals.
  This can potentially lead to erroneous statistics of the total
  charge, leading to sub-optimal or misinformed state and federal
  policy decisions. Missing data rates for total charge are at 2.08\%
  for the NIS 2012 dataset. However, the Michigan SID dataset missing
  data rate for total charge is significantly higher at, 19.79\%,
  underscoring the episodic problems of much higher impact for missing
  data in certain communities. The total number of samples is
  7,296,968.

In this section we contrast the accuracy performance of the multilevel 
Kriging/BLUP method with the recommended imputation algorithms in
HCUP report \#2015-01, which include Predicted Mean
Matching (PMM). We make further comparison with 
more traditional methods such as K-Nearest Neighbors (KNN) and KNN
regression. In particular, we test the accuracy of the various methods
on the above-mentioned total charge variable, containing the highest missing
data rate. The different methods where tested on the 2013 NIS dataset, which
was available to us for analysis. Our calculations show the overall missing data rate is 
2\% for total charge. The multi-level
representation leads to significant computational benefits when
computing $\bbeta$ and the prediction $\hat Y(\bx_0)$ in equation
\eqref{KrigBLUP}.

There is a wealth of publications that attempt to address computational
costs of the above approaches.  Most of these approaches are challenged by stringent
a priori assumptions on the statistical properties of data
(\cite{Sun2012, Sun2015, Stein2013, Stein2012,
  Stein2004,Furrer2011,Furrer2006,Anitescu2012}).

Recently, from the computational mathematics community, a promising
hierarchical matrix approach has been developed by \cite{Litvinenko2019}
to accurately compress covariance matrices, leading to
significant speed-ups. However, the application is restricted to zero-mean data without a component trend.  Another promising approach is
based on the pivoted Cholesky decomposition
developed in \cite{Liu2019}. Nonetheless, if the condition number of the covariance
matrix is large, as happens in practice often, these numerical
methods will have difficulty solving the Kriging problem with any accuracy.
A common ad hoc technique to improve the condition number involves the
use of a so-called nugget; however, it is known that this leads to numerical inaccuracies.

\emph{Ill-conditioned matrices cannot be inverted with accuracy
  (see \cite{Jabbari})}.  \crb{Furthermore, a covariance matrix can in general be
  large and unable to reside in computer memory}.  Thus the
  solution of the GLS would require inverting the covariance matrix $p$
  times using an iterative method, with $p$ the number of columns of the design matrix.

The multi-level approach developed in \cite{Castrillon2013,
  Castrillon2016a, Castrillon2020}, avoids forming the covariance
matrix $\bC$, by transforming the problem into a multi-level form with
significantly smaller condition numbers. In particular, for the BLUP
problem the transformation is one-to-one and onto. \emph{This implies
  that the solution for the multi-level form exactly solves the
  original BLUP problem \eqref{KrigBLUP}.}  Although this appears
impossible due to the ill-conditioning and accuracy issue mentioned above, it can be shown that the prediction is a solution to a
constrained optimization problem (e.g. unbiased constraint).
Constructing a multi-level basis that spans a complementary
constrained space leads to well-conditioned and accurate numerical
algorithms. Furthermore, a single matrix inversion is all that is
required.

\section{Methods: Multilevel approach}
\label{multilevelapproach}

We describe the main ideas of the multilevel approach are
  used to tackle the above-mentioned numerical challenges.  The details of this method
 developed by \cite{Castrillon2020} can be involved for the reader not well
 versed in advanced numerical analysis; here We present a simplified exposition.

Let $\mcP^{p}(\bbS)$ be the span of the columns of the design matrix
$\bX$. Suppose that there exist orthogonal projections $\bL : \R^N
\rightarrow \mcP^{p}(\bbS)$ and $\bW : \R^N \rightarrow
\mcP^{p}(\bbS)^{\perp}$, where $\mcP^{p}(\bbS)^{\perp}$ is the
orthogonal complement of $\mcP^{p}(\bbS)$.  The operators $\bL$ and
$\bW$ are constructed efficiently with an oct or binary kd-tree as
shown in Theorem \ref{Multilevelapproach:theo2}.

\begin{remark}
\jrev{The operator $\bL$ and $\bW$ are constructed from a multilevel decomposition of the location of predictors.  This process is somewhat elaborate and the reader is referred to \cite{Castrillon2016a} and \cite{Castrillon2020} for all of the details. However, for the exposition in this section it sufficient to know what the properties of the operators $\bL$ and $\bW$ are.}
\end{remark}

\begin{theorem} 
\label{Multilevelapproach:theo2}
Suppose that we have a kd-tree representation, with $t$ levels, of 
all the observation locations in $\bbS$. Then:
\begin{enumerate}[i)]
    \item The linear operators $\bL$ and $\bW$ can be constructed in
      $\mcO(Nt)$ computational steps and memory.
    \item The linear operators $\bL$ and $\bW$ have at most $\mcO(Nt)$ 
    non-zero elements.
    \item The operator $\left[
\begin{array}{c}
\bW \
\bL
\end{array}
\right]$ is orthogonal.
\end{enumerate}
\end{theorem}
\begin{proof} See \cite{Castrillon2016a,Castrillon2020}.  
\end{proof}

\begin{remark}
For most practical datasets $\bbS$, the number of levels of the
kd-tree $t$ is $ \approx \log_{2}{N}$.
\end{remark}
Letting $\bY_{\bW}: = \bW \bY$, from equation
\eqref{Introduction:vectormodel} it follows that $\bY_{\bW} = {\bf W}
({\bX \bbeta}+ {\boldsymbol \varepsilon}) = {\bf W{\boldsymbol
    \varepsilon}}$. Note that trend ${\bX} \bbeta$ is
filtered from the data ${\bf Y}$. The new log-likelihood function for
the estimation of $\btheta$ becomes
\begin{align}\label{Introduction:multilevelloglikelihood}
    \begin{split}
   \ell_{\bW}(\btheta)
=&-\frac{n}{2}\log(2\pi)-\frac{1}{2}
\log \det\{\bC_{\bW}(\btheta)\}\\
&-\frac{1}{2}\bY_{\bW}\T\bC_{\bW}(\btheta)^{-1}\bY_{\bW},     
    \end{split}
\end{align}
where $\bC_{\bW}(\btheta) = \bW \bC(\btheta) \bW \T$ and
$\bY_{\bW}\sim\mcN_{N-p} (\0, \bW \bC(\btheta) \bW\T)$.  One immediate
consequence is that the  likelihood function is decoupled. Furthermore,
the following theorem shows that
$\bC_{\bW}(\btheta)$ is more stable numerically than
$\bC(\btheta)$.

\begin{theorem} \label{Multilevelapproach:theo1}
If $\kappa(A) \rightarrow \R$ is the condition number of the
matrix $A \in \R^{N \times N}$, then
\[
\kappa(\bC_{\bW}(\btheta)) \leq 
\kappa(\bC(\btheta)).
\]
\end{theorem}
\begin{proof} See \cite{Castrillon2016a,Castrillon2020}.  
\end{proof}


\begin{remark}
Evaluating the likelihood function $\ell_{\bW}(\btheta)$ requires the computation
of $\log \det\{$ $\bC_{\bW}(\btheta)\}$ and solving 
$\bC_{\bW}(\btheta)^{-1}$ $\bY_{\bW}$. This  can be done by constructing a
Cholesky factor (see \cite{Golub1996}) of $\bC_{\bW}$.
From Theorem \ref{Multilevelapproach:theo1}
this is more stable numerically than evaluating $\ell(\btheta)$. Nonetheless,
the computational efficiency can be significantly increased by constructing
a sparse matrix version of $\bC_{W}$, which we refer to as $\tilde \bC_{W}$,
that is close to the full dense matrix. The sparse matrix $\tilde \bC_{W}$
is built using a distance criterion approach (\cite{Castrillon2016a,Castrillon2020}).
A sparse Cholesky factorization can now
be computed. An alternative method for evaluating
$\bC_{\bW}(\btheta)^{-1}$ $\bY_{\bW}$ is an iterative method such as CG  iteration (\cite{Golub1996}). This
is discussed more in detail below. Note that in practice to estimate 
$\btheta$ accurately, it is unnecessary
to compute the Cholesky factor of the entire sparse matrix $\tilde \bC_{W}$, 
but just sparse sub-blocks. The computational burden will be significantly 
reduced (See \cite{Castrillon2016a, Castrillon2020} for details).
\end{remark}

We now show how to construct a multilevel predictor that gives rise to
well conditioned multilevel covariance matrices. As pointed out in our 
problem setup, this is equivalent to a best linear unbiased
predictor but much easier to solve numerically, making it suitable
for missing data problems in large datasets. 
Consider the system of equations


\[
\begin{pmatrix}
 \bC(\btheta)  & \bX  \\
 \bX\T  & \0  \\
\end{pmatrix}
\begin{pmatrix}
 \hat \bgamma \\
 \hat \bbeta  \\
\end{pmatrix}
=
\begin{pmatrix}
 \bY \\
 \0 \\
\end{pmatrix}
\label{Kriging:problem}
\]

In \cite{Nielsen2002} the authros show
that the solution of \eqref{Kriging:problem}
is given by the GLS estimate of $\bbeta$ (equation \eqref{GLSbeta})
and $\hat \bgamma(\btheta) = \bC^{-1}(\btheta)(\bY - \bX \hat 
\bbeta(\btheta))$. The BLUP at the targe point $\bx_0$ is given
by
\begin{equation}
  \hat Y(\bx_0) = \bk(\bx_0)\T\hat \bbeta(\btheta)+\bc(\btheta)\T \hat \bgamma(\btheta).
\label{Kriging}
\end{equation}
Furthermore,  the Mean Squared Error (MSE) at the target 
point $\bx_0$ is given by
\[
1 + 
\tilde{\bu}
\T
(\bX \T 
\bC(\btheta)^{-1} \bX )^{-1}
\tilde{\bu}
-\bc(\btheta)\T\bC^{-1}(\btheta)\bc(\btheta),
\]
where $\tilde{\bu}\T := (\bX \bC^{-1}(\btheta)
\bc(\btheta) - \bk(\bx_0))$.

From \eqref{Kriging:problem} the key observation is
that $\bX\T \hat
\bgamma(\btheta) = \0$, which implies that $\hat{\bgamma} \in \R^{N}
\backslash \mcP^{p}(\mathbb{S})$. Thus it can be  rewritten as
$\hat{\bgamma} = \bW\T \bgamma_{\bW}$ for some $\bgamma_{\bW} \in
\R^{N-p}$. From equation \eqref{Kriging:problem}
rewrite $\bC(\btheta) \hat \bgamma + \bX \hat \bbeta
= \bY$ as
\begin{equation}
\bC(\btheta) \bW\T \bgamma_{\bW} + \bX \hat \bbeta =
       \bY.
\label{Kriging:eqn1}
\end{equation}
Now apply the matrix $\bW$ to equation \eqref{Kriging:eqn1} and obtain
$\bW \{\bC(\btheta) \bW\T \bgamma_{\bW} + \bX \hat \bbeta\} = \bW
\bY.$ Since the columns for $\bX$ are in $\mcP^{p}(\mathbb{S})$,
it follows
$\bW \bX = \0$ and
\begin{equation}
\bC_{\bW}(\btheta)
\bgamma_{\bW} = \bY_{\bW}.
\label{Kriging:eqn2}
\end{equation}
The advantage of this form is that $\bC_{\bW}(\btheta)$ is better conditioned
due to Theorem \ref{Multilevelapproach:theo1} and $\bgamma_{W}$ can 
be solved by applying a numerical inversion algorithm. Next, $\hat \bgamma$ can be computed
by applying the transformation $\hat{\bgamma} = \bW\T \bgamma_{\bW}$. Finally,
the GLS estimate $\hat \bbeta$ can solved from \eqref{Kriging:eqn1} by applying
the least squares 
\begin{equation}
\hat \bbeta = (\bX^{T}\bX)^{-1}\bX^{T}(\bY 
- \bC(\btheta) \hat \bgamma).
\end{equation}
\jrev{In Figure \ref{fig:workflow} a workflow of the imputation method is shown.}

\begin{remark}
It is remarkable that $\hat \bgamma$ can be solved for independently of the
GLS estimate $\hat \bbeta$ and in turn, $\hat \bbeta$ can be solved as a 
least squares problem without the need to invert the covariance matrix 
$\bC(\btheta)$. \jrev{Furthermore the multilevel solution of the BLUP using equation
(5), (9) and (10) leads to the same exact answer as the original BLUP in equation (5). 
However, from Theorem \ref{Multilevelapproach:theo1} the multievel BLUP is numerically
more stable. See Table \ref{numericalresults:table4} for the differences in 
condition numbers for $\bC$ and $\bC_{\bW}$.
This is of particular importance since it is known that if matrices are ill-conditioned
they cannot be inverted (directly or indirectly) with accuracy \cite{Jabbari}. This is the 
reason that a nugget is usually added to the covariance matrix. However, this changes 
the model to make it easier to solve, but does not solve the original BLUP.}
\end{remark}

The linear system of equations \eqref{Kriging:eqn2} can be solved using a
direct or iterative approach. If $N$ is relatively small, a direct method
such as a Cholesky factorization (\cite{Golub1996}) will work well. However, for
large $N$, due to well-conditioning of the matrix $\bC_W(\btheta)$, a CG 
(\cite{Golub1996})
method is a better approach.  Let $\bgamma_{\bW}^n$ be the $n^{th}$ conjugate
gradient estimate of $\bgamma_{\bW}$, where $\bgamma_{\bW}^0$ is the initial
guess. The main cost of the CG method is in computing the matrix vector products
$\bC_{\bW}(\btheta) \bgamma_{\bW}^n$. This can be done in 3 steps as:
\[
\bgamma_{\bW}^n 
\xrightarrow[(1)]{\bW \T \bgamma_{\bW}^n} 
\ba_n \xrightarrow[(2)]{\bC(\btheta) \ba_n}
\bb_n \xrightarrow[(3)]{\bW \bb_n}
\bC_{\bW}(\btheta) \bgamma_{\bW}^n.
\]
\begin{enumerate}[(1)]

\item The first step transforms $\bgamma_{\bW}^n$ into a single level (original) representation
in $\R^{N}$. 

\item The matrix vector product ${\bC(\btheta) \ba_n}$ is computed
  with a summation method. For problems in $\R^2$ and $\R^{3}$ this we
  can achieve this efficiently ($\mcO(N)$ computational cost) using a
  Kernel Independent Fast Multipole Method (KIFMM)
  (\cite{ying2004,Castrillon2016a}) or a Hierarchical Matrix
  (\cite{Litvinenko2019}). For $d > 3$ dimensions the direct approach is
  used with a cost of $\mcO(N^2)$.

\item For the last step, 
$\bb_n$ is  transformed to a multilevel representation and the matrix vector 
product $\bC_{\bW}(\btheta) \bgamma_{\bW}^n$ is obtained.
\end{enumerate}

\begin{remark}
A preconditioner $\bP_{\bW}$ can be used to speed up the convergence rate
of the CG method, and the system of equations 
\begin{equation}
\bP_{\bW}^{-1}\bC_{\bW}(\btheta)
\bgamma_{\bW} = \bP_{\bW}^{-1} \bY_{\bW}
\label{Kriging:eqn3}
\end{equation}
is solved
instead of \eqref{Kriging:eqn2}.
For the multilevel method, $\bP_{\bW}$ can be constructed  using the diagonal
entries of $\bC_{\bW}$. 
Note that it is possible that $\bC_{\bW}(\btheta)$ will have small condition numbers. 
If this is the case no preconditioner is used.
\end{remark}

\begin{remark}
Given $k$ CG iterations, the total computational cost for computing 
$\hat{\bY}(\bx_0)$ from \eqref{Kriging} using the multilevel approach 
is $\mcO(p^{3} + (k + 1)N^{\alpha} + 2Nt)$. For 2 and 3 dimensional problems, 
the parameter $\alpha$ is $1$ with the use of the KIFMM method. For higher dimensions, 
a direct approach is used, and thus $\alpha = 2$. \crb{The residual error for the CG
method decays exponentially with respect to $k$ and  at a rate that is a function of
the condition number. Small condition numbers lead to fast convergence, 
see \cite{Golub1996} for details.} 
\end{remark}

\begin{remark}
  The multilevel method is implemented in MATLAB \cite{matlab2016} and
    C/C++.  More details can be found in the paper by
    \cite{Castrillon2020}. However, for this paper we have further
    optimized the code which now runs at least twice as fast.
  \end{remark}

\begin{figure}
\begin{tikzpicture}
  \begin{scope}
   
      \node[block] at (2,0) (data)
    { \begin{tabular}{c} 
     Data 
       \end{tabular}};
       
       \node[block] at (2,2) (training)
    { \begin{tabular}{c} 
     Training \\  $(\bx_T,\by_T)$
     
       \end{tabular}};

    \node[block] at (2,-2) (val)
    { \begin{tabular}{c} 
     Missing values
     \\  predictors $\bx_0$
     
       \end{tabular}};

        \node[block] at (5,2) (Multi)
    { \begin{tabular}{c} 
     Multilevel
     \\  $\bL$, $\bW$
     
       \end{tabular}};

    \node[block] at (8,2) (Estimation)
  { \begin{tabular}{c} 
     Estimation
     \\  $l_{\bW}(\hat \btheta)$
       \end{tabular}};

        \node[block] at (5,2) (Multi)
    { \begin{tabular}{c} 
     Multilevel
     \\  $\bL$, $\bW$
     
       \end{tabular}};

    \node[block] at (6.5,-2) (BLUP)
  { \begin{tabular}{c} 
     BLUP (Imp.)
     $\hat Y(\bx_0) = $ \\$\bk(\bx_0)\T\hat \bbeta(\hat \btheta)+\bc(\hat \btheta)\T \hat \bgamma(\hat \btheta)$
       \end{tabular}};
       \draw[vecArrow] (data.north) -- (training.south);
       \draw[vecArrow] (data.south) -- (val.north);
       \draw[vecArrow] (training.east) -- (Multi.west); 
       \draw[vecArrow] (Multi.east) -- (Estimation.west);
       \draw[vecArrow] (Estimation.south) -- (8,-1.45);
       \draw[vecArrow] (Multi.south) -- (5,-1.45);
        \draw[vecArrow] (val.east) -- (BLUP.west);
       \draw[vecArrow] (3,1.5) -- (4.5,-1.45); 
  \end{scope}
\end{tikzpicture}
    \caption{\jrev{Multilevel Kriging/BLUP flowchart. \textbf{Training:} The
    data is split into the predictors $\bx_T$ and the observation
    data $\by_T$ corresponding to the variable that will be imputed.
    \textbf{Multilevel:} The multilevel operators are constructed $(\bL,\bW)$.
    \textbf{Estimation:} The coefficients $\hat \btheta$ of the covariance coefficients
    are estimated. \textbf{BLUP (Imputation):} Given the multilevel operators $(\bL,\bW)$,
    the covariance coefficients $\hat \btheta$ and the predictors for the missing
    values $\bx_0$ imputing the missing variable.}
    }
    \label{fig:workflow}%
\end{figure}

 \section{Experiments and results}
\label{section:ER}

As discussed in the introduction, missing data in the NIS and SID
  datasets form an important problem underscored in the HCUP report
  \#2015-01 (\cite{Houchens2015}). The report highlights, the problem
  of missing data for discharge information (total charge) in rural
  hospitals with potential consequences involving erroneous statistics
  and consequently possibly sub-optimal and even misinformed state and
  federal policy decisions. The missing data rates for total charge
  are at 2.08\% for the NIS 2012 data. The Michigan SID data has a
  total charge missing data rate significantly higher at 19.79\%. We
  test the multilevel approach on the NIS 2013 dataset.  The NIS 2013
  missing data rate for total charge was 2.00 \%.

In this section we contrast the accuracy performance of the multilevel 
Kriging/BLUP method against recommended imputation algorithms in
HCUP report \#2015-01, including Predicted Mean
Matching and Predicted Posterior Distribution methods. We make further comparison with
more traditional methods such as K-Nearest Neighbors (KNN) and KNN
regression. In particular, we test the accuracy of the various methods
on the total charge variable, with its highest missing
data rate.

The computational and accuracy performance of the multi-level
Kriging method is analyzed with
the following choice of 
Mat\'{e}rn covariance function
\[
\phi(r, \rho, \nu):
=\frac{1}{\Gamma(\nu)2^{\nu-1}} \left(
\sqrt{2\nu}\frac{r}{\rho} \right)^{\nu} K_{\nu} \left(
\sqrt{2\nu}\frac{r}{\rho} \right),
\]
with $\Gamma$ the gamma function, $\nu > 0$, $\infty > \rho > 0$,
and where $K_{\nu}$ is the modified Bessel function of the second kind. 
The parameter $\nu$ controls the shape of the Mat\'{e}rn kernel and
$\rho$ is the length correlation. Thus for this
case $\btheta = (\nu,\rho)$, and the stochastic optimization approach seeks
the estimate of $\btheta$ that best explains the data.

To demonstrate the numerical efficiency of the multi-level method, we
generate a series of random observation nodes on a n-sphere
$\mcS_{d-1} := \{\bx \in \R^{d}\,\,|\,\,\|\bx\|_{2} = 1 \}$ with
dimension $d$.  \crb{We create a series of nested sets of nodes
  $\bS_{1}^{d} \subset \dots \subset \bS_{7}^{d}$ that vary with $N =
  2000, 4000$ to $N = 128,000$ in size.  The final set $\bS^d_{7}$
  contains 128,000 randomly selected points on the n-sphere
  $\mcS_{d-1}$. For the covariates we choose the first $d-1$
  dimensions. In other words, forming a matrix of node coordinates
  ($N$) by the number of dimensions $d$, we pick the first $d-1$
  columns as our covariate nodes. The last dimension (column) is
  chosen to be the observations.}  The polynomial basis chosen for the
design matrix $\bX$ is Total Degree (\textbf{TD}) with maximum degree
$w$.

The imputation performance of the multi-level Kriging method is tested
on the National Inpatient Sample (NIS) datasets (\cite{NIS2012}),
Healthcare Cost and Utilization Project (HCUP), Agency for Healthcare
Research and Quality with the 2013 data set.  Among all 190 variables
in this dataset, {\tt totchg} (total charge), 
as, the most problematic, is
a good candidate to test the performance of the multi-level method.
The variables {\tt npr} (number of procedures), {\tt ndx} (number of
diagnoses), {\tt los} (length of stay) and {\tt age} are used as predictors.
Note that as an experimental comparison we also test {\tt los} as a candidate 
response variable, though this would not be necessary in practice 
since its  missing data rate is 0.004 \% in the NIS 2013
dataset.
We extract from the
NIS 2013 data matrix these five variables and remove any incomplete rows.
To test the imputation performance of the multi-level Kriging method,
$N$ rows are selected at random for $N = \{ 2,000$; $5,000$; $10,000$;
$50,000$; $100,000$ \}.

\begin{table*}[!htpb]
\centering
(a)
  $\btheta = (\nu,\rho) = (5/4,10)$, $d = 20$, $w = 3$ ($p = 1771$) \\
\renewcommand{\arraystretch}{1}
\begin{tabular} { r c c c  c r r r r r}
  \multicolumn{1}{c}{$N$} & $\kappa (\bC)$ & $\kappa (\bC_{\bW})$ &
 itr($\bC$)
  & 
 itr($\bC_{\bW}$) &  MB(s) & Itr(s) & Total(s) &  Eff$_{\bgamma,\bbeta}$ \\
  \hline
   \rowcolor{blue!10} 
 16,000  & $5 \times 10^{7}$ &  6  & 178 & 10  &  52  &      56   & 109    &  31,520  \\
 32,000  & $1 \times 10^{8}$ & 10  & 237 & 13  & 125  &     276   & 403    &  32,290  \\
  \rowcolor{blue!10} 
 64,000  & $3 \times 10^{8}$ & 17  & 303 & 16  &  288 &     1405  & 1,695  &  33,540  \\
\end{tabular}\\
\bigskip
(b) $\btheta = (\nu,\rho) = (5/4,10)$, $d = 25$, $w = 2$ ($p = 351$) \\
\renewcommand{\arraystretch}{1}
\begin{tabular} { r c c c c r r r r}
  \multicolumn{1}{c}{$N$} & $\kappa (\bC)$ & $\kappa (\bC_{\bW})$
  & itr($\bC$)
  & itr($\bC_{\bW}$) & MB(s) & Itr(s) & Total(s) & Eff$_{\bgamma,\bbeta}$ \\
  \hline
   \rowcolor{blue!10} 
 16,000  &  $3  \times 10^{7}$  &  17  &  139 
 &  17 &  7 &    89 &  98  &  3,510  \\
 32,000  & $ 8 \times 10^{7}$   &  33  &  183  &  22 & 16 &    462 & 480     &  2,920  \\
  \rowcolor{blue!10}   
 64,000  &  $ 2 \times 10^{8}$  &   64    &  231    &  29 & 35  & 2,400 & 2,437 &   2,800  \\ 
 128,000  &  -                  &  -      & -       &  38 &  78  & 12,644 &  12,724  &  - \\
\end{tabular}
 \caption{Multilevel BLUP Kriging results for the n-sphere data set
    with $d = 20$ and $d = 25$ dimensions, \textbf{TD} design matrix
    of degree $w$, and Mat\'{e}rn covariance function with parameters
    $(\nu,\rho)$.  (a) Computational wall-clock times for solving the
    Kriging prediction for $d = 20$ and $\btheta = (5/4,10)$.  Due to
    the direct method to compute the matrix vector product, the
    computational burden increases somewhat faster than quadratic.
    However, compared to the single level iterative approach it is
    $\approx 33,540$ faster for $N = 64,000$ observations. (b) Kriging
    prediction for $d = 25$ and $\btheta = (5/4,10)$.  For $N =
    64,000$ observations the efficiency of the multilevel BLUP is
    about 2,796 times faster for the same accuracy.} 
\label{numericalresults:table4}
\end{table*}

The error performance is measured using the relative
root-mean-square error (rMSE), mean absolute percentage error (MAPE),
and \jrev{the mean of the} log of the accuracy Ratio (lnQ). The relative RMSE represents
the sample standard deviation of the differences between predicted and
observed values normalized by the mean of the square of the observed
values. The MAPE corresponds to the averaging the ratio of differences
between predicted values and observed values to observed
values; here there is a bias towards small predictions. lnQ
overcomes this issue by using an accuracy measure based on the ratio of
the predicted to actual value.

The Kriging  predictor is compared with other methods such as the
Generalized Least Squares (GLS), k-nearest neighbors (KNN) and KNN
regression. Comparisons are also made with the
following four well known imputation methods: PMM (predicted mean
matching), PPD (posterior prediction distribution), BEM (bootstrapping
EM), and DA (data augmentation). \jrev{
Furthermore, we obtain results for the  Discrimitative Deep Learning (DDL)
from the AutoML library \emph{autokeras} software package \cite{Jin2019}. The model is optimized by setting the number 
of trials to 50 and the number of \emph{epochs} to 50 also \cite{jager2021benchmark}.}

\begin{table*}[!htpb]
\centering
\renewcommand{\arraystretch}{1}
\begin {tabular}{ l c c c  }
 \multicolumn{4}{c}{(a) {\tt totchg} imputation} \\
 \multicolumn{4}{c}{} \\
 Methods &rMSE &MAPE & lnQ\\
 \hline
 \rowcolor{blue!10} 
 PMM   &0.864 & 1.235 &1.00\\
 PPD &  0.869 & 3.378 &1.779\\
 \rowcolor{blue!10} 
 BEM &0.869 &3.317 &1.745\\
 DA    & 0.867 & 3.449 &1.787\\
 \rowcolor{blue!10}
 Kriging    &0.535 & 0.861& 0.492\\
\end{tabular}
\hspace{1.25cm}
\renewcommand{\arraystretch}{1}
\begin{tabular}{ l c c c  }
 \multicolumn{4}{c}{(b) {\tt totchg} Imputation (with log trans.)} \\
 \multicolumn{4}{c}{} \\
 Methods &rMSE &MAPE & lnQ \\
 \hline
 \rowcolor{blue!10} 
 PMM   &0.802& 1.102 & 0.888\\
 PPD &  0.967&1.117& 0.924\\
 \rowcolor{blue!10} 
 BEM &1.092 &1.171& 0.943\\
 DA    & 0.968 &1.192& 0.935\\
 \rowcolor{blue!10}
 Kriging    &0.545& 0.653& 0.418\\
\end{tabular}
\caption{Imputation performance comparison of Kriging/BLUP, PMM, PPD, BEM
  and DA for the {\tt totchg} variable with $N = 100,000$ data points.
  (a) Imputation performance without transformation. The Kriging
  approach clearly outperforms the state of the art methods.  (b)
  Imputation performance with log transformation. The Kriging also
  outperforms the state of the art methods. Furthermore, other methods
  such as DA degrade with the transformation.  }
\label{performance:table1}
\end{table*}

Kriging/BLUP provides 38\% reduction in error  for rMSE, 75\% for MAPE and 72\% for mean lnQ compared to PPD (see Table \ref{performance:table1}). \jrev{Similar performance  is also achieved compared to PPM,  DA and BEM for rMSE, but significantly better for MAPE and lnQ. We will analyze this in 
more detail in this section.} Our error rates are significantly lower than the state of the art methods
recommended by HCUP report\#2015-01, with up to a 75\% reduction. Indeed, this can have a strong impact on funding as an example of policy decision-making. 
As an example, if for half of a group of rural hospitals the total charge is missing, 
mean estimates could be significantly off under recommended methods, with poor funding and related policies as a consequence. In particular, our numerical results
show that MAPE errors for PPD, BEM, DA and can be more than 390\% greater than the multilevel method, with a figure of 140\% for 
PMM.

The numerical performance of the multilevel approach is tested on the
datasets $\bS^{d}_{k}$ for $k = 4,\dots,7$, $d = 20$ and $d = 25$
dimensional problems.  Since $d > 3$, a fast summation (convolution)
method such as the KIFMM is unavailable. Each matrix-vector product of
the conjugate gradient iterations is computed with the direct approach
using a combination of the Graphics Processing Unit (GPU, Nvidia GTX
970) and a single i7-3770 CPU @ 3.40GHz processor.

\begin{figure*}[htbp]
\centering
\begin{tikzpicture}
  \begin{scope}
    \node at (-2.5,-0.75) [] {{\includegraphics[scale = 0.41, trim = 11mm 109mm 120mm 0mm, clip]{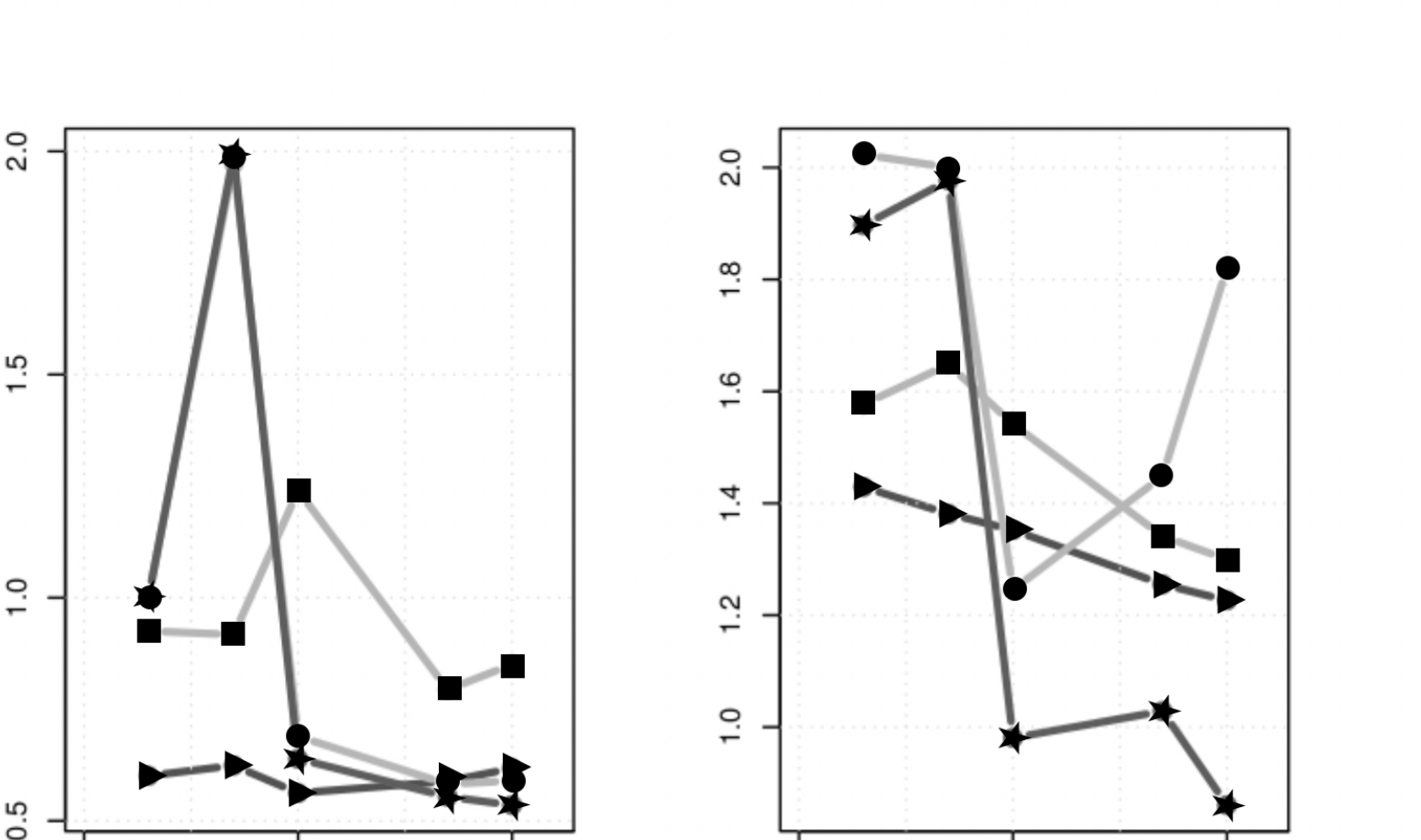} }};
    \node at (-5.75,-4.35)  {$10^{3}$ \hspace{8.5mm} $10^{4}$ \hspace{9mm} $10^{5}$};
    \node at (-.13,-4.35)  {$10^{3}$ \hspace{8.5mm} $10^{4}$ \hspace{9mm} $10^{5}$};
    \node at (-5.70,1.9)  {rMSE};
    \node at (-0,1.9)  {MAPE};
  \end{scope}
\end{tikzpicture}
\begin{tikzpicture}
  \begin{scope}
    \node at (5.4,-0.75) [] {{\includegraphics[scale = 0.41, trim = 270mm 109mm 10mm 0mm, clip]{Errorrate.pdf} }};
    \node at (6,0.55)  {kNN};
    \node at (6.2,-1)  {GLS};
    \node at (6.6,-2)  {kNN-R};
    \node at (6.5,-3.25) {\textbf{Kriging}};
    \node at (5.65,-4.35)  {$10^{3}$ \hspace{8.5mm} $10^{4}$ \hspace{9mm} $10^{5}$};
    \node at (5.55,1.9)  {lnQ};    
  \end{scope}
\end{tikzpicture}
    \caption{Prediction error comparison ({\tt totchg}) for the 2013
      NIS Dataset with respect to the number of data points $N$, where
      90\% are used for training and 10\% for validation.}
    \label{fig:performance2}%
\end{figure*}

We test the performance of BLUP only since the multi-level estimation
computational burden is almost negligible in comparison (See
\cite{Castrillon2020} for more detail).  In Table
\ref{numericalresults:table4} (a) and (b) numerical results for
computing the BLUP parameters $\hat \bgamma$ and $\hat \bbeta$ for $d
= 20$ and $\btheta =(5/4,10)$ are shown. The CG relative residual
tolerance accuracy is set to $tol = 10^{-3}$, itr($\bA$) is the number
of CG iterations needed to achieve $tol$ residual for any matrix
$\bA$. MB(s) is the wall-clock time in seconds needed to compute the
multilevel basis. Itr(s) is similarly time needed to
solve for $\hat \bgamma$ using the CG method. Total(s) is total
time needed to solve for $\hat \bbeta$ and $\hat
\bgamma$.  An efficiency comparison between the cost of computing
$\hat \bbeta$ and $\hat \bgamma$ with the original covariance matrix
$\bC$ and the multilevel approach is given by Eff$_{\bgamma,\bbeta}$.

We first notice that the condition number for $\bC$ is large
($\kappa(\bC) \approx 10^{8}$) even for relatively small problems.
This has several numerical stability implications including a sever
downgrade for maximal accuracy using any numerical inversion
algorithm. A single precision computation would lead to erroneous
results. Using a double precision computation can ameliorate the
accuracy problem, but still exhibit slow convergence.  In comparison
$\kappa(\bC_{\bW}) \approx 20$ is significantly smaller leading to a
stable and fast matrix inversion algorithm.

Compared to the traditional iterative approach using the covariance
matrix $\bC$ (single level representation), \emph{the multilevel
  method is thousands to tens of thousands times faster for the same
  accuracy}.  This can be observed in Table
\ref{numericalresults:table4} (a) and (b).  The source of this
efficiency is due to: i) The number of iterations for convergence to
the same tolerance accuracy is significantly smaller. ii) The
multilevel approach only needs one iterative matrix inversion, in
comparison to $p$ iterative matrix inversions needed for the
traditional single level representation approach. This is relevant for
large problems where the matrix cannot reside in memory.


We can now test the accuracy of the Kriging method. In particular we
test the performance of the multi-level Kriging method with respect to
the following regression problems: i) {\tt los} $\sim$ {\tt totchg} +
{\tt npr} + {\tt ndx}, ii) {\tt totchg} $\sim$ {\tt los} + {\tt npr} +
{\tt ndx} + {\tt age} and iii) log({\tt totchg}) $\sim$ log({\tt los})
+ log({\tt npr}) + {\tt ndx} + {\tt age} (normalized).

\begin{figure*}[htbp]
\centering
\begin{tikzpicture}
  \begin{scope}
    \node at (0,-0.75) [] {\includegraphics[width = 2in]
      {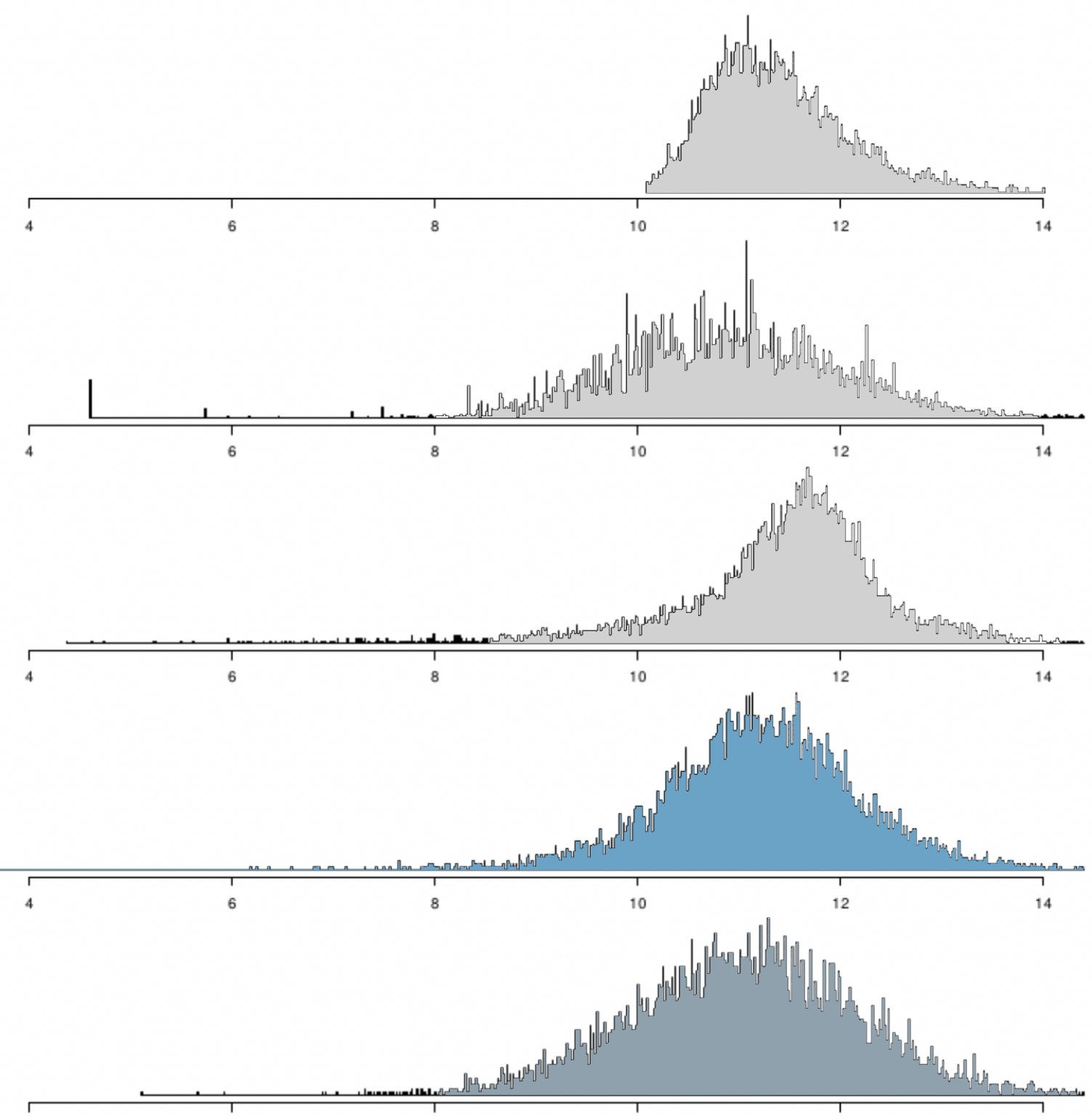} 
      };      
    \node at (-1,1.5)  {kNN-R};
    \node at (-1,0.4)  {kNN};
    \node at (-1,-0.7)  {GLS};
    \node at (-1,-1.75) {\textcolor{black!80}{\textbf{Kriging}}};
    \node at (-1,-2.75) {\textcolor{black!80}{\textbf{Validation}}};
    \node at (6,1)  {\renewcommand{\arraystretch}{1}
\begin{tabular}{ l c c c  }
 Methods &rMSE &MAPE & lnQ\\
 \hline
    \rowcolor{blue!10} KNN(reg)   & 0.618 &1.227 &0.721\\
 KNN&   0.850& 1.300 &1.328\\
   \rowcolor{blue!10}
 GLS & 0.590 &1.821 & 1.011\\
 \jrev{DDL} & \jrev{0.566} & \jrev{0.926} & \jrev{0.637}\\
 \rowcolor{blue!10}
Kriging    &0.535 & 0.861& 0.492\\
\end{tabular}    
    };
    
    \node at (6,-0.6)  {(b) {\tt totchg} imputation};
    
   \node at (6,-2.75)  {\renewcommand{\arraystretch}{1}
   \begin{tabular}{ l c c c  }
Methods & rMSE & MAPE & lnQ \\
 \hline
 \rowcolor{blue!10} 
 KNN(reg)   & 0.677 & 1.021& 0.521\\
 KNN&   0.768& 0.729& 0.583\\
 \rowcolor{blue!10} 
 GLS &0.748 &1.688 &1.898\\
 Kriging    &0.662& 1.189& 0.997\\
\end{tabular}
   };
     \node at (6,-4) 
    {(c) {\tt los} imputation};

    \node at (0,-4)  {(a) $\log({\tt totchg})$ population histogram};

  \end{scope}
\end{tikzpicture}

\caption{(a) Population histogram statistical comparison of kNN-R,
  kNN, GLS and Kriging with respect to the validation data set for
  90,000 training and 10,000 validation datasets ($N =
  100,000$). Notice that Kriging more faithfully reproduces the
  population statistics of the validation total charge data set. This
  is the advantage of the unbiased constrained in the stochastic
  optimization. Note that PMM, PPD, BEM and DA methods also give
  similar results to kNN-R. (b) Total charge ({\tt totchg}) imputation
  statistical errors comparisons. Kriging provides the best imputation
  performance for all error measures.  (c) Length of stay ({\tt los})
  imputation statistical errors comparisons. For {\tt los}, in general
  Kriging performs well.}
    \label{fig:performance}
\end{figure*}

\begin{figure*}[htpb]
\centering
   \includegraphics[scale = 0.45, trim = 50mm 60mm 40mm 60mm, clip]{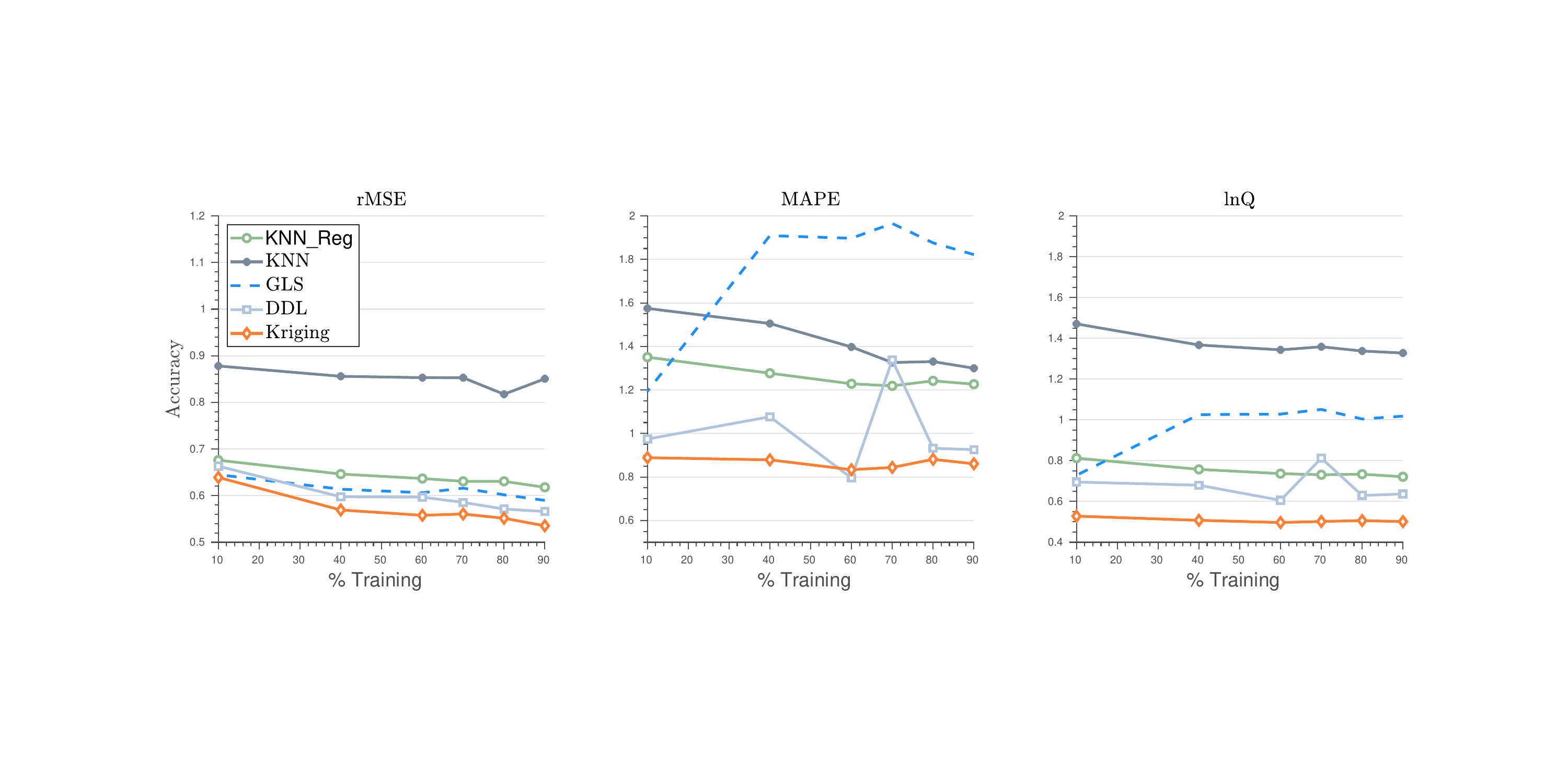} 
   \caption{\jrev{Performance comparison for imputation of the total charge variable among Kriging/BLUP, KNN-Reg,  KNN, GLS and DDL for different training/validation proportions of the data. On the horizontal axis we have the percentage proportion for the training dataset. The vertical axis corresponds to the rMSE, MAPE and mean lnQ metrics. As observed for all the metrics rMSE, MAPE and
   mean lnQ the Kriging/BLUP method produces in almost all cases the best results.}} 
   \label{fig:performance3}
\end{figure*}

\begin{enumerate}[i)]

\item \textbf{{\tt totchg} $\sim$ {\tt los} + {\tt npr} + {\tt ndx} +
  {\tt age}.}  For predicting total charge, we employ {\tt los}, {\tt
  npr}, {\tt ndx} and {\tt age} as predictors for datasets of size $N
  = 2,000$ to $N = 100,000$ with a 90\% training and 10\% validation
  split.  This is a 4 dimensional problem, so we cannot use a fast
  summation method. In Figure \ref{fig:performance2} Kriging/BLUP is
  compared to kNN-R, kNN and GLS for the prediction on the validation
  set.  Observe that as the number of observations increases, Kriging
  outperforms all the other methods. Furthermore from Figure
  \ref{fig:performance} (a) the general shape of the population,
  including mean and variance, of the validation dataset is well
  captured by Kriging, but significantly degrades for kNN-R, KNN and
  GLS. The same phenomenon was also observed for PMM, PPD, BEM and
  DA. In addition, from the table in Figure \ref{fig:performance} (b)
  we observe that Kriging outperforms \jrev{GLS, kNN and DDL} consistently in all
  three measures of accuracy. In Table \ref{performance:table1} (a) we
  observe that Kriging outperforms the 
  imputation packages such as PMM, PPD, BEM and DA for $N = 100,000$
  data points with the 90\%-10\% training/validation split.

\jrev{In Figure \ref{fig:performance3}, we can observe a comparison of performance among different methods for the {\tt totchg} (total charge) variable: Kriging/BLUP, KNN-Reg, KNN, GLS, and DDL. This comparison is conducted across varying training/validation proportions of the dataset (from (10\%/90\%)  to (90\%/10\%) of the data. The horizontal axis depicts the percentage of the dataset used for training, while the vertical axis represents metrics such as rMSE, MAPE, and mean lnQ. Notably, the Kriging/BLUP method consistently outperforms the other methods across all metrics (rMSE, MAPE, and mean lnQ), with its superiority evident in nearly all scenarios.}

\item \textbf{{\tt los} $\sim$ {\tt totchg} + {\tt npr} + {\tt ndx}.}
  For the experiments of predicting length of stay, {\tt totchg}, {\tt
    npr} and {\tt ndx} are used as predictors due to the high
  correlation with {\tt los} and run the simulations for datasets
  (training set plus testing set) of size $N = 100,000$. Note that
  since this is a three dimensional problem, the Kriging multi-level
  code is significantly faster due to the application of the
  KIFMM. From Figure \ref{fig:performance} (c) the result of the
  experiment shows that the Kriging was not always the best predictor
  consistently. However, it still appears to outperform most of the other
  methods. Intuitively, in this case, this could be due to the
  violation of the Gaussian assumption of the linear model.
  

\item log({\tt totchg}) $\sim$ log({\tt los}) + log({\tt npr})
  + {\tt ndx} + {\tt age}. (normalized)

For this experiment the same dataset as in i) is used. However, a log
transformation is and normalization step is applied.  From Table
performance:table1 (b) it is observed that the accuracy of
traditional imputations methods improves. Although the accuracy of the
Kriging method improves somewhat, it still outperforms all
others. This indicates that the proposed Kriging method has the
capacity of handling raw and rough models when traditional methods
tend to fail.

\end{enumerate}

\section{Conclusion}
In this paper we introduce novel techniques from Computational
Applied Mathematics to solve large
scale statistical problems. In particular, the problem of imputation
is solved with the new multi-level Kriging method. Due to the
numerical and stability problems associated with the stochastic
optimization Kriging method, until recently this had limited
applicability to imputation for large datasets. Due to the
introduction of multi-level methods from the CAM community, many of
these limitations have been resolved. Our results show that the
multi-level Kriging method is computationally feasible, stable
numerically, accurate and mathematically principled. In particular, it
is shown that the multilevel BLUP is exact and significantly
outperforms current state-of-the-art methods.  Furthermore, it is
robust and applies to a large class of missing data problems
such as massive medical records.

Multiple imputation is an important strategy for quantifying the
  uncertainty of predictions. There are many methods such as
  bootstrapping used to created multiple realizations of data. These
  realizations are used to quantify the variances of predictions. Such
  methods can also be used to create multiple realizations for the
  multilevel Kriging/BLUP approach. However, the extension is not
  trivial. To more faithfully reproduce realizations of the data, the
  bootstrapping approach needs to take into account the Cholesky
  decomposition (see \cite{Golub1996}) of the covariance matrix $\bC$,
  which is very difficult since the matrix is large and
  ill-conditioned. However, alternatively we can use a Karhunen
  Lo\'{e}ve (KL) expansion (see \cite{Castrillon2022}) to create
  multiple realizations with the Mat\'{e}rn covariance function from
  the Gaussian process representation of the data. This would involve
  computing the eigenstructure of the covariance function, which is
  significantly more stable to compute even if the matrix is
  ill-conditioned.  Moreover, by using a Kernel Independent Fast
  Multipole (KIFMM) approach \cite{ying2004} the computation of the
  eigenstructure in principle can be relatively fast for problems in
  $\R^3$.  An advantage of the KL expansion is that it does not
  involve inversion of the covariance matrix. However, for higher
  dimensional problems, such as predicting the total charge missing
  data involving 4 dimensions, there are no known fast summation
  methods such as the KIFMM. It is still possible to compute the
  eigenstructure for a relatively large dataset on a powerful Graphics
  Processing Unit (GPU).

Alternatively, there exists a set of linear equations for the BLUP 
that solve for the Mean Square Error (MSE) of the prediction. 
This involves inverting the covariance matrix $\bC$; thus for large datasets
it can be numerically unstable and intractable. \cite{Castrillon2016a} show that there
exists a formulation for the multilevel approach that is significantly faster and numerically
stable. However, it can still be intractable for estimating large numbers of missing data. We
shall investigate these approaches in more detail in a future publication. 

\jrev{Finally, we have not addressed the problem of imputation for categorical data, which
is part of the HCUP dataset.
 Our approach can potentially deal with the categorical data by treating it as numerical and defining a cutoff. For example this is what is done with Support Vector Machines (SVMs). The question that arises is what choice of cutoff do we use ? We will deal with these types of variables in a future publication.}

\ifCLASSOPTIONcompsoc
  \section*{Acknowledgments}
\else
  \section*{Acknowledgment}
\fi

We appreciate the help and advice from Bindu Kalesian, in particular,
for the access to the NIS dataset. Futhermore, the feedback from
Karen Kafadar has been invaluable.
This material is based upon work
supported by the National Science Foundation under Grant No. 1736392.

\ifCLASSOPTIONcaptionsoff
  \newpage
\fi



%


%

\begin{IEEEbiography}[{\includegraphics[width=1in,height=1.25in,clip,keepaspectratio]{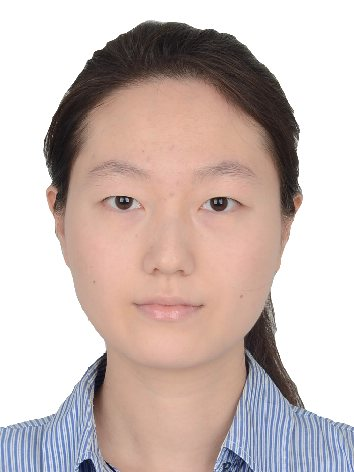}}]{Wenrui Li} obtained a BS degree in statistics from Shandong University, a MS degree in statistics from the University of Washington, and a PhD degree in statistics from Boston University. She is currently a postdoc in the Department of Biostatistics, Epidemiology and Informatics at the University of Pennsylvania. Her research is focused on causal inference, network analysis, epidemic parameter estimation and knowledge-guided statistical learning.
\end{IEEEbiography}

\begin{IEEEbiography}[{\includegraphics[width=1in,height=1.25in,clip,keepaspectratio]{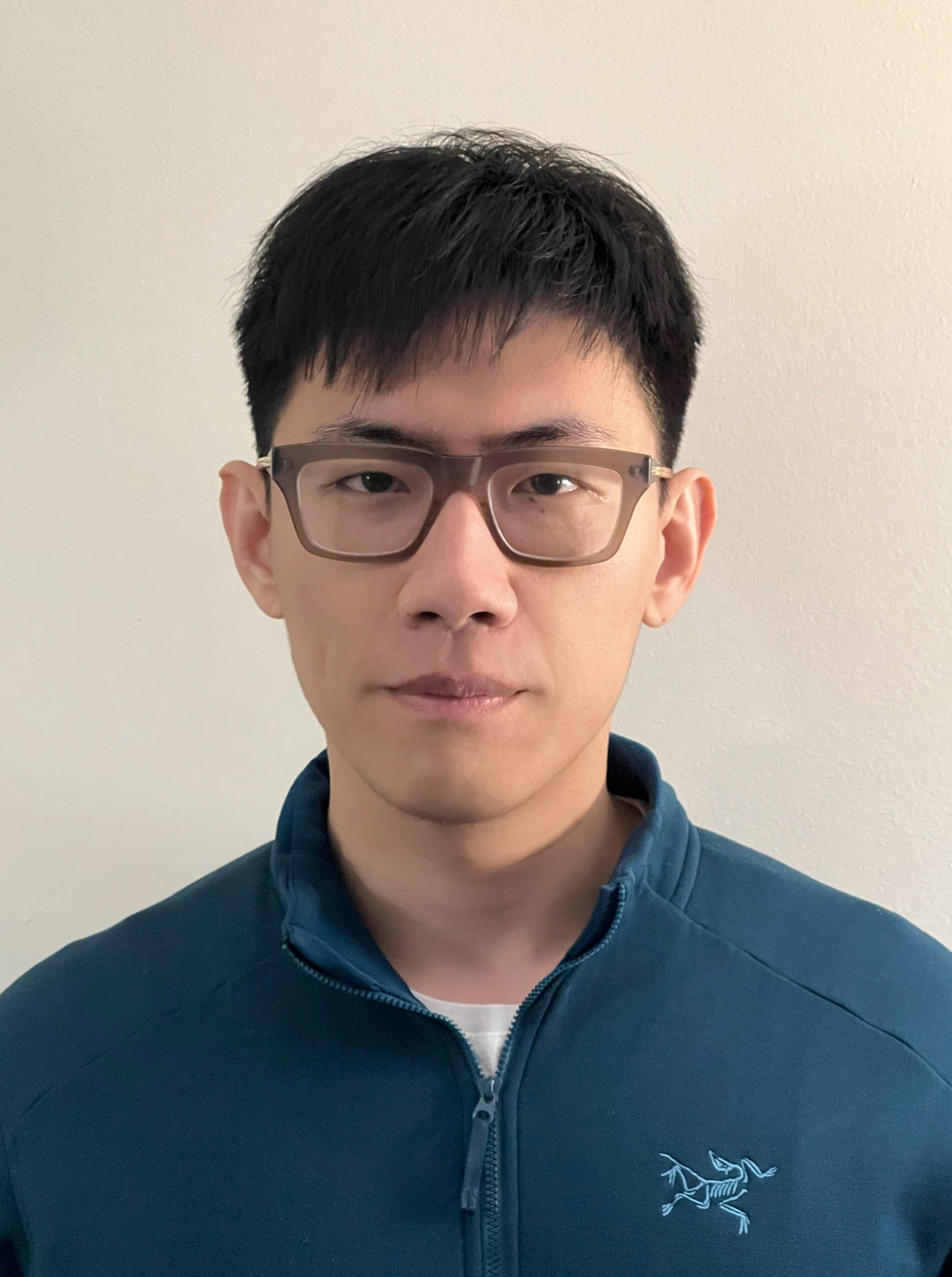}}]{Xiaoyu Wang} obtained a BS degree in mathematics from University of Science and Technology of China, a MA degree in statistics from Boston University. He is currently pursuing his Ph.D. degree in statistics at Boston University. His research is focused on the asymptotic behavior of stochastic optimization algorithm.
 \end{IEEEbiography} 

\begin{IEEEbiography}[{\includegraphics[width=1in,height=1.25in,clip,keepaspectratio]{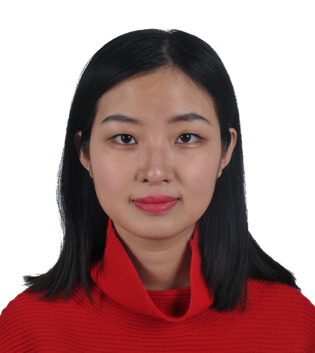}}]{Yuetian Sun} obtained a BS degree in statistics from Beijing Normal University, a MA degree in statistics from Boston University. She is currently a senior data consultant at ClearView Healthcare Partners. Her work is focused on leveraging variance healthcare data sources (i.e. Claims, EMR...) to provide marketing strategies and actionable insights on physician targeting, patient journey and treatment landscape and using advanced data science and statistical technologies like web scraping, text mining, conjoint analysis to conduct KOL and influence mapping, market simulation, payer segmentation.
\end{IEEEbiography}

\begin{IEEEbiography}[{\includegraphics[width=1in,height=1.25in,clip,keepaspectratio]{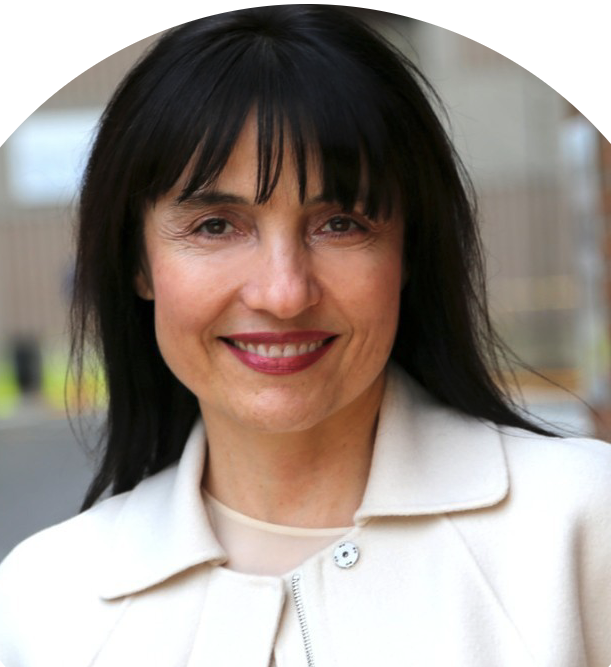}}]{Snezana Milanovic} received a M.Sc. degree in neurosciences and a M.D. degree in University of Belgrade, Serbia. She did postdoctoral training in University of Belgrade, International Reference Laboratory for
Neuroimmunomodulation, State University of New York, Max Planck Institute for Experimental Medicine, Central Institute of Mental Health, Mannheim, Albert Einstein College of Medicine, Yale University, Harvard School of Medicine, Massachusetts General Hospital, and Beth Israel Deaconess Medical Center. She is currently a senior director at Sunovion Pharmaceuticals. 
 
\end{IEEEbiography} 

\begin{IEEEbiography}[{\includegraphics[width=1in,height=1.25in,clip,keepaspectratio]{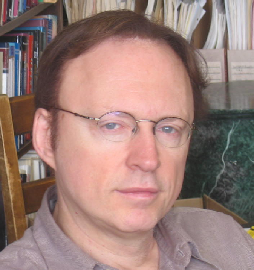}}]{Mark Kon} obtained Bachelor’s degrees in Mathematics,
Physics, and Psychology from Cornell
University, a PhD in Mathematics from MIT. He
is a professor of Mathematics and Statistics
at Boston University. He is affiliated with the
Quantum Information Group, the Bioinformatics
Program and the Computational Neuroscience
Program. He has had appointments at Columbia
University as Assistant and Associate Professor
(Computer Science, Mathematics), as well as at
Harvard and at MIT. He has published approximately
100 articles in mathematical physics, mathematics and statistics, 
computational biology, and computational neuroscience, including two
books. His recent research and applications interests involve quantum
probability and information, statistics, machine learning, computational
biology, computational neuroscience, and complexity.
 
\end{IEEEbiography} 

\begin{IEEEbiography}[{\includegraphics[width=1in,height=1.25in,clip,keepaspectratio]{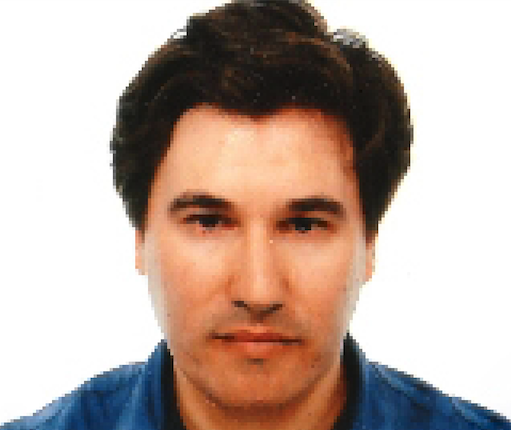}}]{Julio Enrique Castrill\'on-Cand\'as} received the MS and Ph.D. degrees in electrical engineering and computer science from the Massachusetts Institute of Technology (MIT), Cambridge. He is currently faculty in the department of Mathematics and Statistics at Boston University. His area of expertise is in Uncertainty Quantification (PDEs, non-linear stochastic networks), large scale computational statistics, functional data analysis and statistical machine learning.
 
\end{IEEEbiography} 





\end{document}